\documentclass[sn-mathphys,Numbered]{sn-jnl}

\usepackage{graphicx}%
\usepackage{multirow}%
\usepackage{amsmath,amssymb,amsfonts}%
\usepackage{amsthm}%
\usepackage{mathrsfs}%
\usepackage[title]{appendix}%
\usepackage{xcolor}%
\usepackage{textcomp}%
\usepackage{manyfoot}%
\usepackage{booktabs}%
\usepackage{algorithm}%
\usepackage{algorithmicx}%
\usepackage{algpseudocode}%
\usepackage{listings}%

\theoremstyle{thmstyleone}%
\theoremstyle{thmstyletwo}%

\theoremstyle{thmstylethree}%

\begin{document}

\title{Multimodal Machine Learning in Image-Based and Clinical Biomedicine: Survey and Prospects}

\author[1]{\fnm{Elisa} \sur{Warner}}\email{elisawa@umich.edu}

\author[1]{\fnm{Joonsang} \sur{Lee}}\email{leejoons@umich.edu}

\author[2]{\fnm{William} \sur{Hsu}}\email{whsu@mednet.ucla.edu}

\author[3]{\fnm{Tanveer} \sur{Syeda-Mahmood}}\email{stf@us.ibm.com}

\author[4]{\fnm{Charles E} \sur{Kahn, Jr.}}\email{ckahn@upenn.edu}

\author[5]{\fnm{Olivier} \sur{Gevaert}}\email{ogevaert@stanford.edu}

\author[1]{\fnm{Arvind} \sur{Rao}}\email{ukarvind@med.umich.edu}

\affil[1]{\orgdiv{Department of Computational Medicine and Bioinformatics}, \orgname{University of Michigan Ann Arbor}, \orgaddress{\street{100 Washtenaw Ave}, \city{Ann Arbor}, \postcode{48109}, \state{MI}, \country{USA}}}

\affil[2]{\orgdiv{Department of Medical \& Imaging Informatics}, \orgname{University of California Los Angeles}, \orgaddress{\street{924 Westwood Blvd Ste 420}, \city{Los Angeles}, \postcode{90024}, \state{CA}, \country{Country}}}

\affil[3]{\orgdiv{Almaden Research Center}, \orgname{IBM}, \orgaddress{\street{650 Harry Rd}, \city{San Jose}, \postcode{95120}, \state{CA}, \country{USA}}}

\affil[4]{\orgdiv{Department of Radiology}, \orgname{University of Pennsylvania}, \orgaddress{\street{3400 Spruce St.}, \city{Philadelphia}, \postcode{19104}, \state{PA}, \country{USA}}}

\affil[5]{\orgdiv{Center for Biomedical Informatics Research}, \orgname{Stanford}, \orgaddress{\street{1265 Welch Road}, \city{Stanford}, \postcode{94305}, \state{CA}, \country{USA}}}


\abstract{Machine learning (ML) applications in medical artificial intelligence (AI) systems have shifted from traditional and statistical methods to increasing application of deep learning models. This survey navigates the current landscape of multimodal ML, focusing on its profound impact on medical image analysis and clinical decision support systems. Emphasizing challenges and innovations in addressing multimodal representation, fusion, translation, alignment, and co-learning, the paper explores the transformative potential of multimodal models for clinical predictions. It also highlights the need for principled assessments and practical implementation of such models, bringing attention to the dynamics between decision support systems and healthcare providers and personnel. Despite advancements, challenges such as data biases and the scarcity of ``big data" in many biomedical domains persist. We conclude with a discussion on principled innovation and collaborative efforts to further the mission of seamless integration of multimodal ML models into biomedical practice.}

\keywords{machine learning, multimodal, representation, fusion, translation, alignment, co-learning, artificial intelligence, data integration}

\maketitle


\section{Introduction}\label{sec1}

\par Machine learning (ML), the process of leveraging algorithms and optimization to infer strategies for solving learning  tasks, has enabled some of the greatest developments in artificial intelligence (AI) in the last decade, enabling the automated segmentation or class identification of images, the ability to answer nearly any text-based question, and the ability to generate images never seen before. In biomedical research, many of these ML models are quickly being applied to medical images and decision support systems in conjunction with a significant shift from traditional and statistical methods to increasing application of deep learning models. At the same time, the importance of both plentiful and well-curated data has become better understood, coinciding as of the time of writing this article with the incredible premise of OpenAI's ChatGPT and GPT-4 engines as well as other generative AI models which are trained on a vast, well-curated, and diverse array of content from across the internet \cite{openai2023gpt4}. 

\par As more data has become available, a wider selection of datasets containing more than one modality has also enabled growth in the multimodal research sphere. Multimodal data is intrinsic to biomedical research and clinical care. While data belonging to a single modality can be conceptualized  as a way in which something is perceived or captured in the world into an abstract digitized representation such as a waveform or image, multimodal data aggregates multiple modalities and thus consists of several intrinsically different representation spaces (and potentially even different data geometries). Computed tomography (CT) and positron emission tomography (PET) are specific examples of single imaging modalities, while magnetic resonance imaging (MRI) is an example itself of multimodal data, as its component sequences T1-weighted, T2-weighted, and fluid-attenuated inversion recovery (FLAIR) can each be considered their own unique modalities, since each of the MR sequences measure some different biophysical or biological property. Laboratory blood tests, patient demographics, electrocardiogram (ECG) and genetic expression values are also common modalities in clinical decision models. This work discusses unique ways that differences between modalities have been addressed and mitigated to improve accuracy of AI models in similar ways to which a human would naturally be able to re-calibrate to these differences.

\par There is conceptual value to building multimodal models. Outside of the biomedical sphere, many have already witnessed the sheer power of multimodal AI in text-to-image generators such as DALL$\cdot$E 2, DALL$\cdot$E 3 or Midjourney \cite{Ramesh2022HierarchicalTI,Ramesh2023,Oppenlaender2022}, some of whose artful creations have won competitions competing against humans \cite{Metz_2022}. In the biomedical sphere, multimodal models provide potentially more robust and generalizable AI predictions as well as a more holistic approach to diagnosis or prognosis of patients, akin to a more human-like approach to medicine. While a plethora of biomedical AI publications based on unimodal data exist, fewer multimodal models exist due to cost and availability constraints of obtaining multimodal data. However, since patient imaging and lab measurements are decreasing in cost and increasing in availability, the case for building multimodal biomedical AI is becoming increasingly compelling. 
\begin{figure}[h]
\centering
    \includegraphics[scale=0.70]{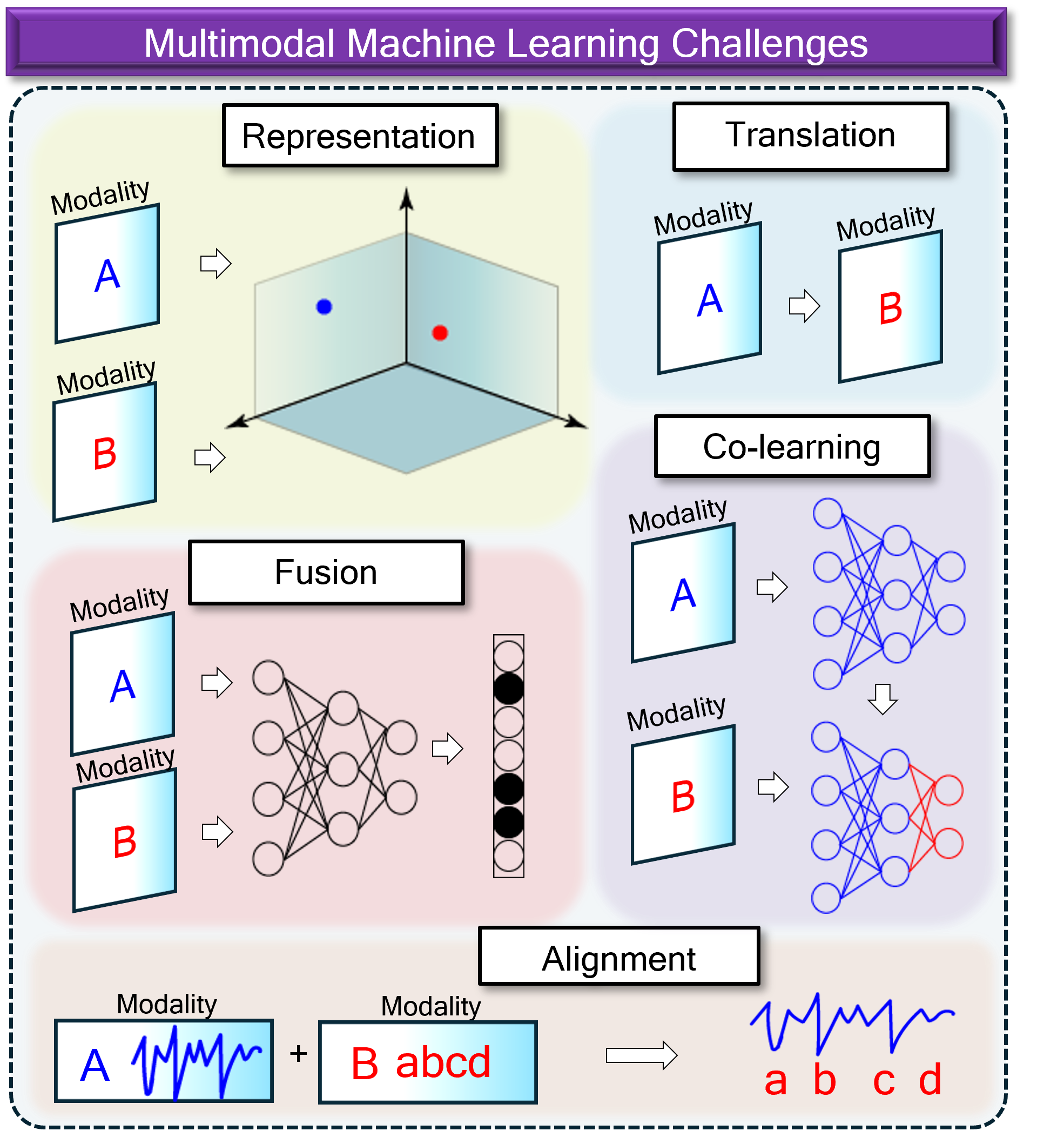}
    \caption{Challenges in multimodal learning: 1) Representation, which concerns how multiple modalities will be geometrically represented and how to represent intrinsic relationships between them; 2) Fusion, the challenge of combining multiple modalities into a predictive model; 3) Translation, involving the mapping of one modality to another; 4) Alignment, which attempts to align two separate modalities spatially or temporally; and 5) Co-learning, which involves using one modality to assist the learning of another modality.}
    \label{fig:figure1}
\end{figure}

\par With the emergence of readily-available multimodal data comes new challenges and responsibilities for those who use them. The survey and taxonomy from \cite{baltrusaitis2017multimodal} presents an organized description of these new challenges, which can be summarized in Figure \ref{fig:figure1}: 1) Representation, 2) Fusion, 3) Alignment, 4) Translation, 5) Co-learning. \textbf{Representation} often condenses a single modality such as audio or an image to a machine-readable data structure such as a vector, matrix, tensor object, or other geometric form, and is concerned with ways to combine more than one modality into the same representation space. Good multimodal representations are constructed in ways in which relationships and context can be preserved between modalities. Multimodal \textbf{fusion} relates to the challenge of how to properly combine multimodal data into a predictive model. In multimodal \textbf{alignment}, models attempt to automatically align one modality to another. In a simple case, models could be constructed to align PPG signals taken at a 60Hz sampling frequency with a 240Hz ECG signal. In a more challenging case, video of colonoscopy could be aligned to an image representing the camera’s location in the colon. Multimodal \textbf{translation} consists of mapping one modality to another. For example, several popular natural language processing (NLP) models attempt to map an image to a description of the image, switching from the imaging domain to a text domain. In translational medicine, image-to-image translation tends to be the most common method, whereby one easily-obtained imaging domain such as CT is converted to a harder-to-obtain domain such as T1-weighted MRI. Lastly, multimodal \textbf{co-learning} involves the practice of transferring knowledge learned from one modality to a model or data from a different modality.

\par In this paper, we use the taxonomical framework from \cite{baltrusaitis2017multimodal} to survey current methods which address each of the five challenges of multimodal learning with a novel focus on addressing these challenges in medical image-based clinical decision support. The aim of this work is to introduce both current and new approaches for addressing each multimodal challenge. We conclude with a discussion on the future of AI in biomedicine and what steps we anticipate could further progress in the field.


\section{Multimodal Learning in Medical Applications}\label{sec3}
\par
\begin{figure}
    \centering
    \includegraphics[width=\textwidth]{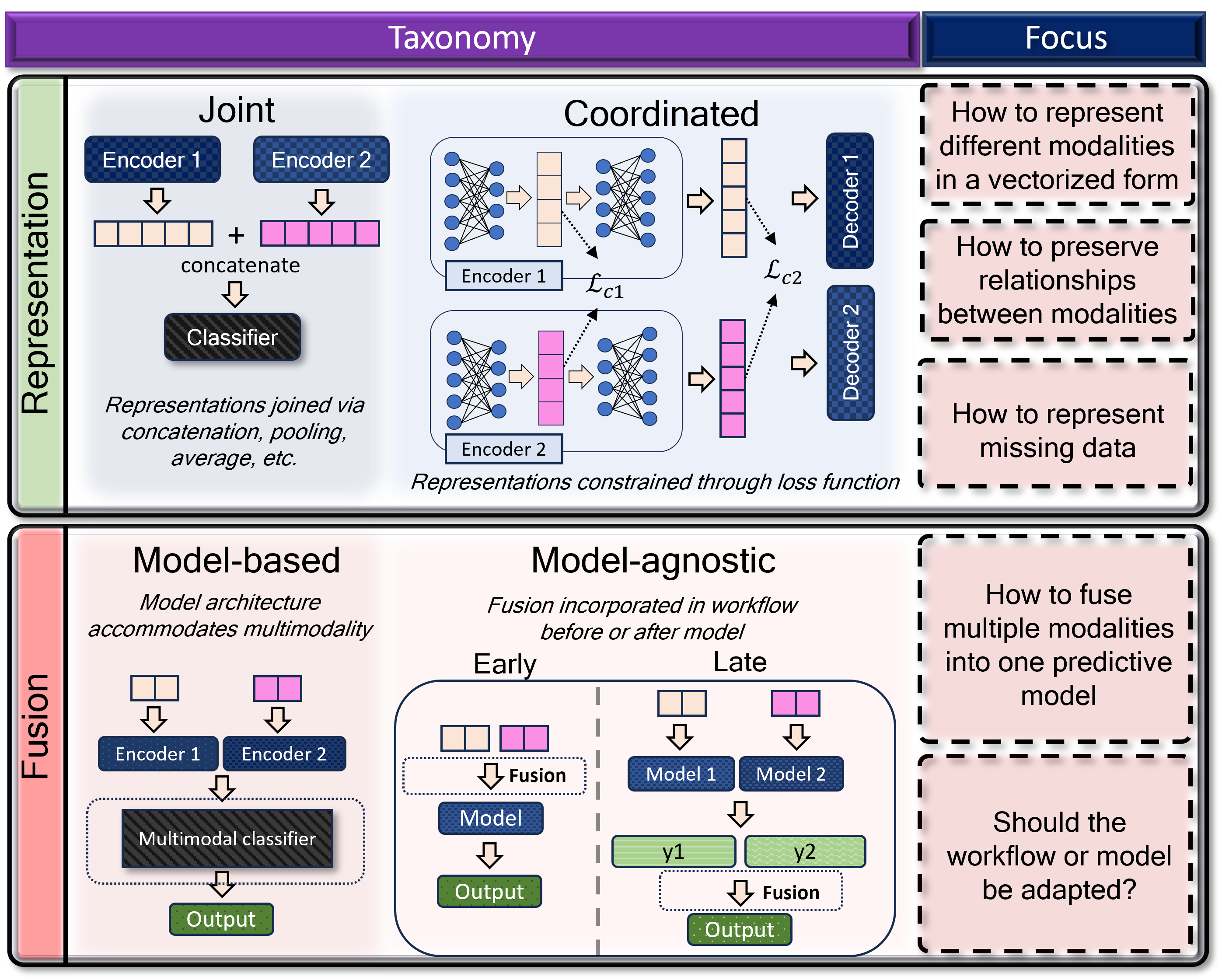}
    \caption{A graphical representation of the taxonomical sublevels of multimodal representation and fusion, and the focus of each challenge. Multimodal representation can be categorized into whether the representations are joined into a single vector (\textit{joint}) or separately constructed to be influenced by each other (\textit{coordinated}). Multimodal fusion can be distinguished by whether a model is uniquely constructed to fuse the modalities (\textit{model-based}), or whether fusion occurs before or after the model step (\textit{model-agnostic}).}
    \label{fig:figure2}
\end{figure}

\begin{figure}[h!]
    \centering
    \includegraphics[width=\textwidth]{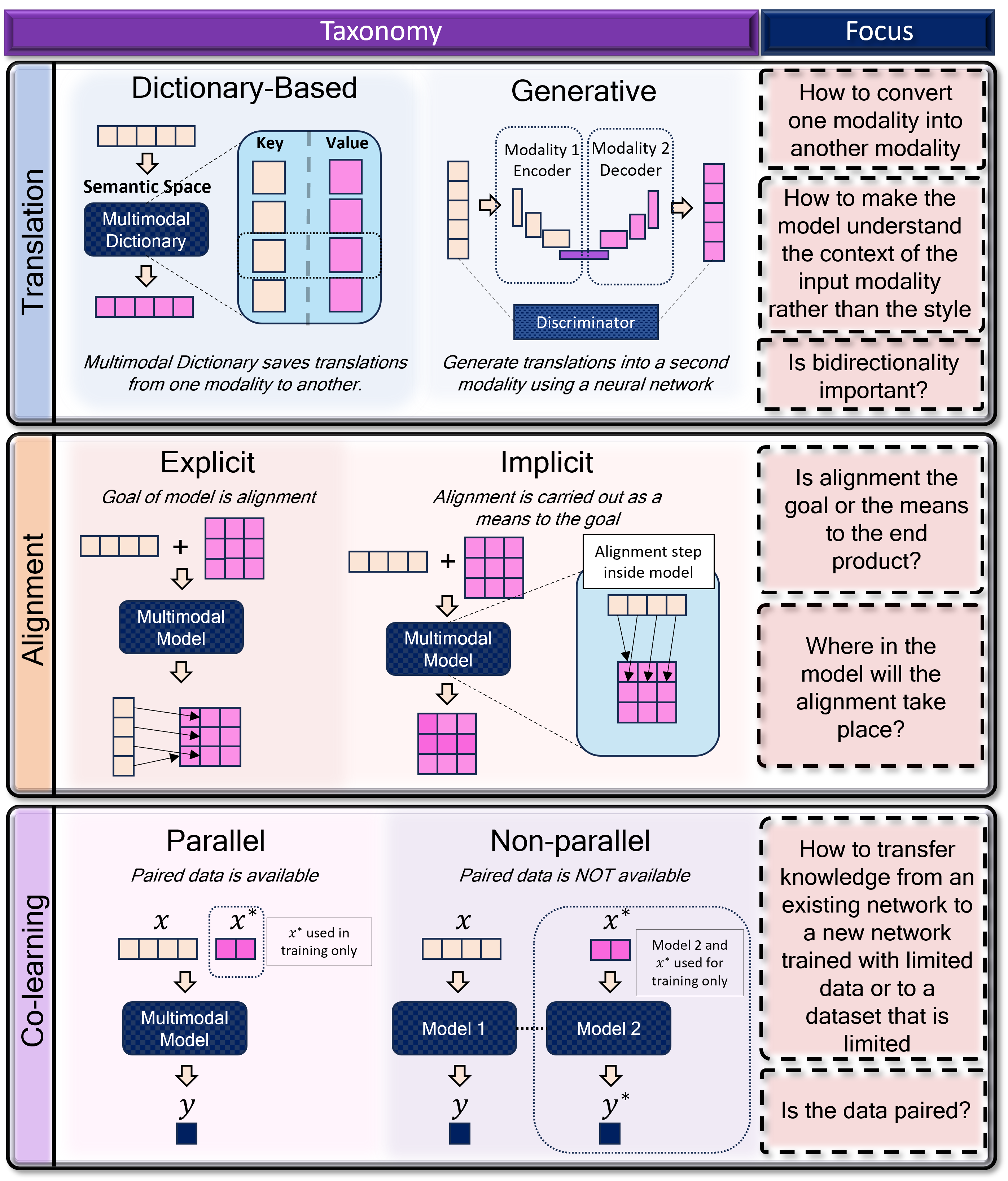}
    \caption{A graphical representation of the taxonomical sublevels of multimodal translation, alignment and co-learning, and the focus of each challenge. In \textbf{translation}, models are distinguished based on whether they require use of a dictionary to save associations between modalities (\textit{dictionary-based}), or if the associations are learned in a multimodal network (\textit{generative}). In \textbf{alignment}, distinction is made depending on the \textit{purpose} of the alignment, whether as the goal (\textit{explicit}) or as an intermediate step towards the goal output (\textit{implicit}). In \textbf{co-learning}, a distinction is made between the use of \textit{parallel} (paired) multimodal data, or \textit{non-parallel} (unpaired) multimodal data. In co-learning models, one of the modalities is only used in training but does not appear in testing.}
    \label{fig:figure3}
\end{figure}

  In the following section, we reintroduce the five common challenges in multimodal ML addressed in Section 1 and discuss modern approaches to each challenge as applied to image-based biomedicine. The taxonomical subcategories of Representation and Fusion are summarized in Figure \ref{fig:figure2}, while those for Translation, Alignment and Co-learning are summarized in Figure \ref{fig:figure3}. A table of relevant works by the challenge addressed and data types used are given in Table 1.

\subsection{Representation}\label{subsec2}
\par Representation in machine learning typically entails the challenge of transferring contextual knowledge of a complex entity such as an image or sound to a mathematically-interpretable or machine-readable format such as a vector or a matrix. Prior to the rise of deep learning, features were engineered in images using techniques such as the aforementioned Scale-Invariant Feature Transform (SIFT) transforms or through methods such as edge detection. Features in audio or other waveform signals such as ECG could be extracted utilizing wavelets or Fourier transform to isolate latent properties of signals and then quantitative values could be derived from morphological patterns in the extracted signal. Multimodal representation challenges venture a step further, consisting of the ability to translate similarities and differences from one modality’s representation to another modality’s representation. For example, when representing both medical text and CT images, if the vector representations for “skull” and “brain” in medical \textit{text} are closer than those for “skull” and “pancreas”, then in a good CT representation, such relationships between vector representations of these structures in the \textit{image} should remain preserved. The derivation of “good” representations in multimodal settings have been outlined in Bengio et al \cite{bengio2013} and extended by Srivastava and Salakhutdinov \cite{JMLR:v15:srivastava14b}.

\par It is crucial to acknowledge that representation becomes notably challenging when dealing with more abstract concepts. In a unimodal context, consider the task of crafting representations from an image. Beyond pixel intensities, these representations must encapsulate contextual and semantically-proximate information from the image. A simplistic model may fail to encode context adequately, discerning insufficient distinctions between a foreground and background to represent nuanced visual-semantic concepts. Achieving such subtleties in representations, particularly in abstract contexts, poses increased challenges compared to quantifying similarities and differences in less-nuanced data such as cell counts or gene expression.

Prior to delving into multimodal representations, it is instructive to elucidate strategies for crafting proficient unimodal representations, as multimodal approaches often involve combining or adapting multiple unimodal methods. For images, \textit{pretrained networks} are a common approach for transforming images into good vector representations. Another approach is use of autoencoders, which condense image representations into lower-dimensional context vectors that can be decoded to reconstruct the original image.  \textit{Multimodal autoencoders} have been applied to MRI modalities in \cite{Hamghalam2021} and in this example were also utilized to impute representations for missing modalities.

\par Another approach for multimodal representation could be through the use of \textit{disentanglement networks}, which can separate latent properties of an image into separate vectors. In such cases, an image is given as input and the autoencoder is often split in such a way that two vectors are produced as intermediate pathways, where joining the intermediate vectors should result in the original input. Each intermediate pathway is often constrained by a separate loss function term to encourage separation of each pathway into the desired latent characteristics. In this way, one input image can be represented by two separate vectors, each representing a disjointed characteristic of the image. This disentanglement method has been applied in \cite{jiang2020unified} to separate context in CT and MRI from their style so that one modality can be converted in to the other. It was also applied for a single modality in \cite{Bne2020} to separate ``shape" and ``appearance" representations of an input, which could potentially be applied to different imaging modalities to extract only similar shapes from each.

\par When two or more vectorized modalities are combined into a model, they are typically combined in one of two ways: 1) joint, or 2) coordinated representations. A \textbf{joint representation} is characterized by aggregation of the vectors at some point in the process, whereby vector representations from two separate modalities are joined together into a single vector form through methods such as aggregation, concatenation or summation. Joint representation is both a common and effective strategy for representation; however, a joint strategy such as concatenation is often constricted to serving in situations where both modalities are available at train- and test-time (one exception using Boltzmann Machines can be found in \cite{JMLR:v15:srivastava14b}). If a modality has the potential to be missing, a joint strategy such as aggregation via weighted means could be a better option \cite{Li2021,Chen2020,Zhou2023,Cui2022}. Using mathematical notation from \cite{baltrusaitis2017multimodal}, we can denote joint representations $x_m$ as the following:

\begin{equation}
    x_m=f(x_1,...,x_n)
\end{equation}

\par This denotes that feature vectors $x_i, i =1...n$ are combined in some way through a function $f$ to create a new representation space $x_m$. By the contrary, \textbf{coordinated representations} are represented as the following:

\begin{equation}
    f(x_1)\sim g(x_2),
\end{equation}

whereby a function designed to create representations for one modality may be constrained (represented by $\sim$) by a similar function from another modality, with the assumption that relationships between data points in the first modality should be relatively well-preserved in the second modality.

\par Joint representations tend to be the most common approach to representing two or more modalities together in a model because it is perhaps the most straightforward approach. For example, joining vectorized multimodal data together through concatenation before entering a model tends to be one of the most direct approaches to joint representation. In \cite{vanSonsbeek2020}, for example, chest x-rays are combined with text data from electronic health records in a vectorized form using a pretrained model first. Then, the vectors from each modality are sent individually through two attention-based blocks, then concatenated into a joint feature space to predict a possible cardiovascular disease and generate a free-text “impression” of the condition. Other joint representation models follow simpler methods, simply extracting baseline features from a pretrained model and concatenating them \cite{Daza2020,Yang2020}. 

\par Although coordinated representations have traditionally tended to be more challenging to implement, the convenience of neural network architectural and loss adjustments have resulted in increased traction in publications embodying coordinated representations \cite{Xing2022,Wang2023b,chauhan2020joint,pmlr-v139-radford21a,Zhang2022MLHC,Bhalodia2021}. One of the most notable of these in recent AI approaches is OpenAI's Contrastive Language-Image Pre-Training (CLIP) model, which forms representations for OpenAI's DALL$\cdot$E 2 \cite{pmlr-v139-radford21a,Ramesh2022HierarchicalTI} and uses a contrastive-learning approach to shape both image embeddings of entire images to match text embeddings of entire captions describing those images. The representations learned from CLIP were demonstrated to not only perform well in zero-shot image-to-text or text-to-image models, but also to produce representations that could outpace baseline supervised learning methods. In a biomedical context, similar models abound, including ConVIRT, a predecessor and forerunner for CLIP \cite{Zhang2022MLHC}, and related works \cite{Bhalodia2021}.

\par Coordinated approaches are especially useful in co-learning. In \cite{chauhan2020joint}, which employs a subset of co-learning called privileged information, the geometric forms of each modality are not joined into a single vector representation. Instead, network weights are encouraged to produce similar output vectors for each modality and ultimately the same classifications. This constraint warps the space of chest x-ray representations closer to the space of text representations, with the assumption that this coordinated strategy provides chest x-ray representations more useful information because of the text modality. For more on privileged information, see the Section 2.5 below.

\par In the biomedical sphere, where models are built to prioritize biologically- or clinically-relevant outcomes, quality of representations may often be overlooked or overshadowed by emphasis on optimization of prediction accuracy. However, there is conceptual value in building good multimodal representations. If models are constructed to ensure that similar concepts in different modalities also demonstrate cross-modal similarity, then there is greater confidence that an accurate model is understanding cross-modal relationships. While building good cross-modal representations for indexing images on the Internet like in the CLIP model is a digestible challenge, building similar cross-modal representations for medical data presents a far more formidable challenge due to data paucity. OpenAI's proprietary WebTextImage dataset, used for CLIP, contains 400 million examples, a sample size that is as of yet unheard of for any kind of biomedical imaging data. Until such a dataset is released, bioinformaticians must often rely on the ability to leverage pretrained models and transfer learning strategies for optimal results amidst low resources to leverage big data for good representations on smaller data.


\subsection{Fusion}\label{subsubsec2}
\par Next, we discuss challenges in multimodal fusion. This topic is a natural segue from the discussion of representation because many multimodal representations are subsequently fed into a discriminatory model. Multimodal fusion entails the utilization of methods to combine representations from more than one modality into a classification, regression, or segmentation model. According to \cite{baltrusaitis2017multimodal}, fusion models can be classified into two subcategories: model-agnostic and model-based approaches. The term \textbf{``model-agnostic"} refers to methods for multimodal fusion occurring either before or after the model execution and typically does not involve altering the prediction model itself. Model-agnostic approaches can further be delineated by the stage at which the fusion of modalities occurs, either early in the model (prior to output generation) or late in the model (such as ensemble models, where outputs from multiple models are combined). Additionally, hybrid models, incorporating a blend of both early and late fusion, have been proposed \cite{LloretCarbonell2023}. In contrast, a \textbf{model-based} approach entails special adjustments to the predictive model to ensure it handles each modality uniquely.

\par While model-agnostic methods remain pertinent as useful strategies for multimodal fusion, the overwhelming popularity of neural networks has led to a predominant shift towards model-based methods in recent years. These model-based methods involve innovative loss functions and architectures designed to handle each modality differently. One common model-based fusion strategy is multimodal \textit{multiple instance learning (MIL)}, where multiple context vectors for each modality are generated and subsequently aggregated into a single representation leading to the output classification. The method for aggregation varies across studies, with attention-based approaches, emphasizing specific characteristics of each modality, being a common choice \cite{Li2021,Chen2020,Zhou2023,Cui2022}.

\par The construction of a good model architecture is crucial; however, challenges associated with fusion are often highly contextual, and thus it is important to understand what kinds of data are being utilized in recent models and what problems they try to solve. Most multimodal models understandably incorporate MRI modalities, given that MR images are a natural multimodal medium. Consequently, studies incorporating MRI such as \cite{azcona2020interpretation}, which aims to classify Alzheimer's Disease severity, and \cite{zhou2020m2net}, predicting overall survival in brain tumor patients, exemplify the type of research often prevalent in multimodal image-based clinical application publications. Brain-based ML studies are also popular because of the wide availability of brain images and a strong interest in applying ML models in clinical neuroradiology. However, recent models encompass a myriad of other clinical scenarios predicting lung cancer presence \cite{Daza2020}, segmenting soft tissue sarcomas \cite{Neubauer2020}, classifying breast lesions \cite{habib2020automatic}, and predicting therapy response \cite{Yang2020}, among others, by amalgamating and cross-referencing modalities such as CT images \cite{Daza2020,Neubauer2020}, blood tests \cite{Yang2020}, electronic health record (EHR) data \cite{Yang2020,vanSonsbeek2020,Daza2020}, mammography images \cite{habib2020automatic}, and ultrasound \cite{habib2020automatic}.

\par Multimodal fusion models are emerging as the gold standard for clinical-assisted interventions due to the recognition that diagnosis and prognosis in real-world clinical settings are often multimodal problems. However, these models are not without limitations. For one, standardization across equipment manufacturers or measurement protocols can affect model performance dramatically, and this issue becomes more pronounced as more modalities are incorporated into a model. Second, while fusion models attempt to mimic real-world clinical practice, they face practical challenges that can limit their utility. For instance, physicians may face various roadblocks to obtaining all model input variables due to a lack of permission from insurance companies to perform all needed tests or time constraints. These issues underscore challenges associated with missing modalities, and several studies have attempted to address this concern \cite{LloretCarbonell2023,Zhang2022,Cui2022,Wang2023b,Liu2023}. However, incorporating mechanisms to account for missing modalities in a model is not yet a common practice for most multimodal biomedical models.

\par Lastly, many models are not configured to make predictions that adapt with additional variables. Most models necessitate all variables to be present at the time of operation, meaning that, even if all tests are conducted, the model can only make a decision once all test results have been obtained. In conclusion, in the dynamic and fast-paced environment of hospitals and other care centers, even accurate models may not be suitable for practical use, unless also coupled with mechanisms to handle missing data.

\subsection{Translation}\label{sec4}

\par In multimodal translation, a model is devised to operate as a mapping entity facilitating the transformation from one modality to another. This involves the conversion of input contextual data, such as CT scans, into an alternative contextual data format, such as MRI scans. Before the rise of modern \textbf{generative} methods leveraging multimodal generative adversarial networks (GANs) or diffusion models to generate one modality from another, translation via \textbf{dictionary-based} methods was common, which typically involved a bimodal dictionary whereby a single entry would contain a key belonging to one modality and a corresponding value belonging to the other modality. Dictionary-based translation was uncommon in biomedical research but popular in NLP fields as a way to convert images into text and vice versa \cite{hu2021text,reed2016generative}. The current ascendancy of generative models and the availability of associated coding packages have since catalyzed the growth of innovative translational studies applying generative approaches.

\par Presently, generative models encompass a broad spectrum of potential applications both within and beyond the biomedical domain. Outside the medical sphere, generative models find utility in NLP settings, particularly in text-to-image models like DALL$\cdot$E 2 and Midjourney \cite{hu2021text,Ramesh2022HierarchicalTI,Oppenlaender2022}. Additionally, they are employed in style transfer and other aesthetic computer vision techniques \cite{Huang_2021,cao2018cari,zhu2020unpaired,Liu2018,Palsson_2018_CVPR_Workshops,Zhang2020b}. Within the biomedical realm, generative models have proven efficacious in creating virtual stains for unstained histopathological tissues which would typically undergo hemotoxylin/eosin staining \cite{lu2020data}. Furthermore, these models serve as prominent tools for sample generation \cite{Tseng2017,Piacentino2021,choi2018generating}, particularly in scenarios with limited sample sizes \cite{Chen2021b}. Despite the potential diversity of multimodal translation involving any two modalities, predominant translational efforts in the biomedical realm currently revolve around mapping one imaging modality to another, a paradigm recognized as image-to-image translation.

\par In the contemporary landscape, the integration of simplistic generative models into a clinical context are declining in visibility, while methods employing specialized architectures tailored to the involved modalities are acknowledged for advancing the state-of-the-art in translational work. Within this context, two notable generative translation paradigms for biomedicine are explored: 1) medical image generation models, and 2) segmentation mask models. In the former, many studies attempt to form models that are bidirectional, whereby the intended output can be placed back as input and return an image similar to the original input image. In \cite{Bui2020}, this is resolved by generating deformation fields that map changes in the T1-weighted sequence modality of MRI to the T2-weighted sequence modality. In \cite{Hu2020}, separate forward and backward training processes are defined whereby an encoder representing PET images is utilized to understand the underlying distribution of that modality, allowing for more realistic synthetic generated images from MRI. In one unidirectional example, \cite{shin2020gandalf} modifies a pix2pix conditional GAN network to allow Alzheimer's disease classification to influence synthetic PET image generation. In another interesting example, \cite{Takagi2023} use functional MRI (fMRI) scans and diffusion models to attempt to recreate images of what their subjects had seen. Similarly, diffusion models and magnetoencephalography (MEG) are utilized by Meta for real-time prediction from brain activity of what patients had seen visually \cite{meta}.

\par In the second potential application, image segmentation models in multimodal image-to-image translation must handle additional challenges, creating both a way to generate the output modality as well as a way to segment it. In \cite{jiang2020unified}, a generative model converts CT to MRI segmentation. In a reverse problem to image segmentation, \cite{guo2020lesion} attempts to synthesize multimodal MRI examples of lesions with only a binary lesion mask and a multimodal MRI Atlas. In this study, six CNN-based discriminators are utilized to ensure the authentic appearance of background, brain, and lesion, respectively, in synthesized images.

\par Multimodal translation still remains an exciting but formidable challenge. In NLP and beyond, there have been remarkable successes observed in new image generation within text-to-image models beyond the biomedical sphere. However, the adoption of translation models in biomedical work is evolving at a more measured pace, with applications extending beyond demonstrative feasibility to practical utility remaining limited. Arguments in favor of biomedical translation models are predominantly centered around sample generation for datasets with limited sizes, as the generated medical images must adhere to stringent accuracy requirements. Similar to other challenges in multimodal research, translation models would greatly benefit from training on more expansive and diverse datasets. However, with the increasing digitization of medical records and a refined understanding of de-identification protocols and data sharing rights, the evolution of this field holds considerable promise.


\subsection{Alignment}
\par Multimodal alignment involves aligning two related modalities, often in either a spatial or temporal way. Multimodal alignment can be conducted either \textit{explicitly} as a direct end goal, or \textit{implicitly}, as a means to the end goal, which could be translation or classification of an input. One example of \textbf{explicit alignment} in a biomedical context is image registration. \cite{Leroy2023} highlights one approach to multimodal image registration, where histopathology slides are aligned to their $(x,y,z)$ coordinates in a three-dimensional CT volume. Another is in \cite{Chen2023}, where surgical video was aligned to a text description of what is happening in the video. On the other hand, an example of multimodal \textbf{implicit alignment} could be the temporal alignment of multiple clinical tests to understand a patients progress over time. Such an analysis was conducted in \cite{Yang2020}, where the authors built a customized multi-layer perceptron (MLP) called SimTA to predict response to therapy intervention at a future time step based on results from previous tests and interventions.

\par Literature surrounding alignment has increased since the rise of attention-based models in 2016. The concept of ``attention," which relates to aligning representations in a way that is contextually relevant, is a unimodal alignment paradigm with origins in machine translation and NLP \cite{bahdanau2016neural}. An example use of attention in NLP could be models which try to learn, based on order and word choice of an input sentence, where the subject of the sentence is so that the response can address the input topic. In imaging, attention can be used to highlight important parts of an image that are most likely to contribute to a class prediction. In 2017, Vaswani et al \cite{DBLP:journals/corr/VaswaniSPUJGKP17}, introduced a more sophisticated attention network, named transformers, an encoder-decoder-style architecture based on repeated projection heads where attention learning takes place. Transformers and attention were originally applied to natural language \cite{DBLP:journals/corr/VaswaniSPUJGKP17,bahdanau2016neural,devlin2019bert} but have since been applied to images \cite{parmar2018image,dosovitskiy2021image}, including histopathology slides \cite{Lu2021,Chen2020} and protein prediction \cite{Tunyasuvunakool2021}. \textit{Multimodal transformers} were introduced in 2019, also developed for the natural language community \cite{Tsai2019}. While these multimodal transformers do not contain the same encoder-decoder structure of a traditional transformer architecture, they are hallmarked by crossmodal attention heads, where one modality's sequences intermingle with another modality's sequences.

\par Although typical transformers themselves are not multimodal, they often constitute in multimodal models. The SimTA network mentioned above borrowed the positional encoding property of transformers to align multimodal inputs in time to predict therapy response \cite{Yang2020}. Many models taking advantage of visual transformers (ViT) have also utilized pretrained transformers trained on images for multimodal fusion models. In both the TransBTS \cite{Wang2021trans} and mmFormer models \cite{Zhang2022}, a transformer is utilized on a vector composed of an amalgamation of information from multiple modalities of MRI, which may imply that the transformer attention heads here are aligning information from multiple modalities represented via aggregate latent vectors. The ultimate function of transformers is a form of implicit alignment, and it can be assumed here that this alignment is multimodal.

\par Transformer models have brought a new and largely successful approach to alignment, sparking widespread interest in their applications in biomedical use. Transformers for NLP have also engendered new interest in Large Language Models (LLMs), which are already being applied to biomedical contexts \cite{Tinn2023} and probing new questions about its potential use as a knowledge base for biomedical questions \cite{Sung2021CanLM}.


\subsection{Co-learning}

\par In this last section exploring recent research in multimodal machine learning, the area of co-learning is examined, a field which has recently garnered a strong momentum in both unimodal and multimodal domains. In multimodal co-learning, knowledge learned from one modality is often used to assist learning of a second modality. This first modality which transfers knowledge is often leveraged only at train-time but is not required at test-time. Co-learning is classified in \cite{baltrusaitis2017multimodal} as either parallel or non-parallel. In \textbf{parallel} co-learning, paired samples of modalities which share the same instance are fed into a co-learning model. By contrast, in \textbf{non-parallel} co-learning, both modalities are included in a model but are not required to be paired.
\par
\begin{figure}
    \centering
    \includegraphics[width=\textwidth]{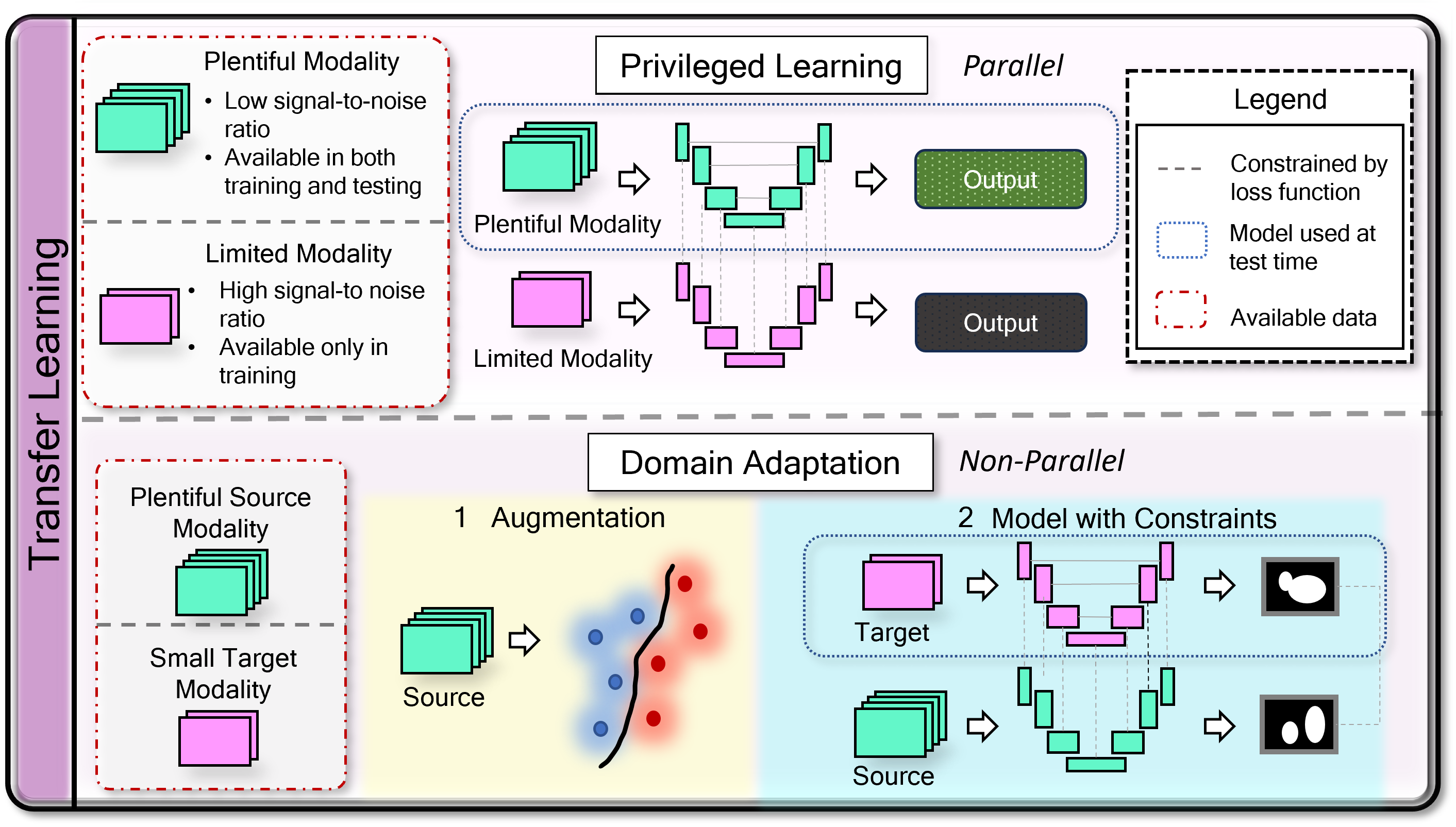}
    \caption{Two types of transfer learning described in this work are privileged learning (top) and domain adaptation (bottom). In privileged learning, a plentiful set consisting of data which is normally of low cost but also low signal-to-noise ratio is available in both training and testing, while a limited gold-standard quality set is used for training only. In this example, the plentiful set is used to train the target model, while the limited set constrains the model parameters to increase the model's ability to associate the low-cost modality with the ground truth. In domain adaptation, there is a target dataset which consists of a few samples and a source dataset consisting of plenty of samples. If the target data is too small to build a reliable model in training, source data can be augmented to make the model more robust. Else, the target model could be trained with few examples, while a second source model is used to help make the target model more generalizable.}
    \label{fig:figure4}
\end{figure}

 While co-learning can embody a variety of topics such as conceptual grounding and zero-shot learning, this work focuses on the use of transfer learning in biomedicine. In \textit{multimodal transfer learning}, a model trained on a higher quality or more plentiful modality is employed to assist in the training of a model designed for a second modality which is often noisier or smaller in sample size. Transfer learning can be conducted in both parallel and non-parallel paradigms. This work focuses on one parallel form of transfer learning called privileged learning, and one non-parallel form of transfer learning called domain adaptation. A visual representation of these approaches be seen in Figure \ref{fig:figure4}.

\subsubsection{Privileged Learning}
\textit{Privileged learning} originates from the mathematician Vladmir Vapnik and his ideas of knowledge transfer with the support vector machine for privileged learning (SVM+) model \cite{Vapnik2009}. The concept of privileged learning introduces the idea that predictions for a low-signal, low-cost modality can be assisted by incorporating a high-signal, high-cost modality (privileged information) in training only, while at test-time only the low-cost modality is needed. In \cite{Vapnik2009}, Vapnik illustrates this concept through the analogy of a teacher (privileged information) distilling knowledge to a student (low-cost modality) before the student takes a test. Although a useful concept, the field is relatively under-explored compared to other areas of co-learning. One challenge to applying privileged learning models was that Vapnik's SVM+ model was one of few available before the widespread use of neural networks. Furthermore, it demands that the modality deemed ``privileged" must confer high accuracy on its own in order to ensure that its contribution to the model is positive. Since then, neural networks have encouraged newer renditions of privileged information models that allow more flexibility of use \cite{Lambert_2018_CVPR,Shaikh2020,Sabeti2021}.
\par Recently, privileged learning has emerged as a growing subset of biomedical literature, and understandably so. Many multimodal models today require health care professionals to gather a slew of patient information and are not trained to handle missing data. Therefore, the ability to minimize the number of required input data while still utilizing the predictive power of multiple modalities can be useful in real-world clinical settings. In \cite{Hu2020b} for example, the authors attempt to train a segmentation network where at train-time the ``teacher network" contains four MR image modalities, but at test-time the ``student network" contains only T1-weighted images, the standard modality used in preoperative neurosurgery and radiology. In \cite{chauhan2020joint}, chest x-rays and written text from their respective radiology reports are used to train a model where only chest x-rays are available at test-time. 

\par In privileged models based on traditional approaches (before deep neural networks), privileged information can be embedded in the model either through an alteration of allowable error (``slack variables" from SVM+) \cite{Vapnik2009}, or through decision trees constructed with non-privileged features to mimic the discriminative ability of privileged features (Random Forest+) \cite{Warner2022,Moradi2016}. In a deep learning model, privileged learning is often achieved through the use of additional loss functions which attempt to constrain latent and output vectors from the non-privileged modality to mimic those from the combined privileged and non-privileged models \cite{Hu2020b,Xing2022}. For example, in \cite{chauhan2020joint}, encoders for each modality are compared and cross entropy loss is calculated for each modality separately. The sum of these allows the chest x-ray network to freely train for only the chest x-ray modality while being constrained through the overall loss function to borrow encoding methods from the text network, which also strives to build an accurate model.

\par While privileged learning models can be applied where data is missing, users should heed caution when applying models in situations where there is systematic bias in reporting. Those who train privileged models without considering subject matter may inadvertently be choosing to include all their complete data in training and their incomplete data in testing. However, in clinical scenarios, data are often incomplete because a patient either did not qualify for a test (perhaps their condition was seen as not ``dire enough" to warrant a test) or their situation was too serious to require a test (for example, a patient in septic shock may not pause to undergo a chest x-ray because they are in the middle of a medical emergency). Therefore, while applying data to highly complex models is a common approach in computer science, the context of the data and potential underlying biases need to be considered first to ensure a practical and well-developed model.

\subsubsection{Domain Adaptation}
\textit{Domain adaptation} has been shown to be useful in biomedical data science applications where a provided dataset may be too small or costly to utilize for more advanced methods such as deep learning, but where a somewhat similar (albeit larger) dataset can be trained by such methods. The smaller dataset for which we want to train the model is called the ``target" dataset and the larger dataset which will be used to assist the model with the learning task and provide better contextualization is called the ``source" dataset. Domain adaptation strategies are often tailored to single modalities such as camera imaging or MRI, where measurements of an observed variable differ based on an instrument's post-processing techniques or acquisition parameters \cite{Xiong2020,varsavsky2020testtime,Yang2020b}. However, the distinct characteristics arising from disparate instruments or acquisition settings can lead to considerable shifts in data distribution and feature representations, mirroring the challenges faced in true multimodal contexts. Therefore, the discussion of uni-modal domain adaptation is a relevant starting point for multimodal domain adaptation, as it covers approaches to mitigate significant deviations within data that may seem similar but are represented differently. Additionally, understanding how to mitigate the impact of such variations helps one to understand ways to construct multimodal machine learning systems that confront similar challenges. We also discuss relevant multimodal domain adaptation approaches in biomedicine, which have typically consisted of applying CT images as a source domain to train an MRI target model or vice versa \cite{Chiou2020, Xue2020, Pei2023, Jafari2022, Dong2022}. 

\par One way to train a model to adapt to different domains is through augmentation of the input data, which ``generalizes" the model to interpret outside of the domain of the original data. In \cite{Xiong2020}, a data augmentation framework for fundus images in diabetic retinopathy (DR) is proposed to offset the domain differences of utilizing different cameras. The authors show that subtracting local average color, blurring, adaptive local contrast enhancement, and a specialized principal component analysis (PCA) strategy can increase both $R^2$ values for age prediction and DR classification area under the receiver operating curve (AUROC) on test sets where either some domain information is known \textit{a priori} and also where no information is known, respectively. In another method which attempts to augment the source domain into more examples in the target style, \cite{Chiou2020} split the source image into latent content and style vectors, using the content vectors in a style-transfer model reminiscent of cycleGAN to feed as examples with the target domain into a segmentation network \cite{Zhu2017}. In other applications, data augmentation for domain generalization may be executed utilizing simpler affine transformations \cite{varsavsky2020testtime}. This demonstrates the utility of data augmentation strategies in more broadly defining decision boundaries where target domains differ from the source.

\par A second strategy for domain adaptation involves constraining neural network functions trained on a target domain by creating loss functions which require alignment with a source domain model. In \cite{varsavsky2020testtime}, a framework for adapting segmentation models at test-time is proposed, whereby an adversarial loss trains a target-based U-Net to be as similar to a source-based U-Net as possible. Then a paired-consistency loss with adversarial examples is utilized to fine-tune the decision boundary to include morphologically similar data points. In a specificially multimodal segmentation-based model, \cite{Xue2020} attempts to create two side-by-side networks, a segmenter and an edge generator, which both encourage the source and target output to be as similar as possible to each other. In the final loss function, the edge generator is used to constrain the segmenter in such a way as to promote better edge consistency in the target domain. In yet another, simpler example, domain adaptation to a target domain is performed in \cite{Hu2021} by taking a network trained on the source domain and simply adjusting the parameters of the batch normalization layer.

\par Domain adaptation in biomedicine can be a common problem where instrument models or parameters change. 
Among multimodal co-learning methods, most networks are constructed as segmentation networks for MRI and CT because they are similar imaging domains, although measuring different things. While CT carries distinct meaning in its pixels (measured in Hounsfield Units), MRI pixel intensities are not standardized and usually require normalization, which could pose challenges to this multimodal problem. Additionally, MRI carries much more detail than CT scans, which necessitates the model to understand contextual boundaries of objects much more than a unimodal case with only CT or MRI.

\section{Discussion} 

The rapidly evolving landscape of artificial intelligence (AI) both within the biomedical field and beyond has posed a substantial challenge in composing this survey. Our aim is to provide the reader with a comprehensive overview of the challenges and contemporary approaches to multimodal machine learning in image-based, clinically relevant biomedicine. However, it is essential to acknowledge that our endeavor cannot be fully comprehensive due to the dynamic nature of the field and the sheer volume of emerging literature within the biomedical domain and its periphery. This robust growth has led to a race among industry and research institutions to integrate the latest cutting-edge models into the healthcare sector, with a particular emphasis on the introduction of ``large language models" (LLMs). In recent years, there has been an emergence of market-level insights into the future of healthcare and machine learning, as exemplified by the incorporation of machine learning models into wearable devices such as the Apple Watch and Fitbit devices for the detection of atrial fibrillation \cite{Perino2021, Lubitz2022}. This begs the question: \textit{where does this transformative journey lead us?}

\par Healthcare professionals and physicians already embrace the concept of multimodal cognitive models in their diagnostic and prognostic practices, signaling that such computer models based on multimodal frameworks are likely to endure within the biomedical landscape. However, for these models to be effectively integrated into clinical settings, they must exhibit flexibility that aligns with the clinical environment. If the ultimate goal is to seamlessly incorporate these AI advancements into clinical practice, a fundamental question arises: how can these models be practically implemented on-site? Presently, most available software tools for clinicians are intended as auxiliary aids, but healthcare professionals have voiced concerns regarding the potential for increased computational workload, alert fatigue, and the limitations imposed by Electronic Health Record (EHR) interfaces \cite{deRuiter2015,Ancker2017}. Therefore, it is paramount to ensure that any additional software introduced into clinical settings serves as an asset rather than a hindrance.

\par Another pertinent issue emerging from these discussions pertains to the dynamics between clinical decision support systems (CDSS) and healthcare providers. What occurs when a computer-generated recommendation contradicts a physician's judgment? This dilemma is not new, as evidenced by a classic case recounted by \cite{Evans1998}, where physicians were granted the choice to either follow or disregard a CDSS for antibiotic prescription. Intriguingly, the group provided with the choice exhibited suboptimal performance compared to both the physician-only and computer-only groups. Consequently, it is unsurprising that some healthcare professionals maintain a cautious approach to computer decision support systems \cite{Adamson2019,Silcox2020}. Questions arise regarding the accountability of physicians if they ignore a correct computer-generated decision and the responsibility of software developers if a physician follows an erroneous computer-generated recommendation.

\par A pivotal ingredient notably under-represented in many CDSS models, which could help alleviate discrepancies between computer-generated and human decisions, is the incorporation of uncertainty quantification, grounded calibration, interpretability and explainability. These factors have been discussed in previous literature, underscoring the critical role of explainability in ensuring the long-term success of CDSS-related endeavors \cite{Reddy2022,Khosravi2022,Kwon2020,Abdar2021}.

\par The domain of multimodal machine learning for medically oriented image-based clinical support has garnered increasing attention in recent years. This interest has been stimulated by advances in computer science architecture and computing hardware, the availability of vast and publicly accessible data, innovative model architectures tailored for limited datasets, and the growing demand for applications in clinical and biomedical contexts. Recent studies have showcased the ability to generate synthetic images in one modality based on another (as outlined in Section 2.3), align multiple modalities (Section 2.4), and transfer latent features from one modality to train another (Section 2.5), among other advancements. These developments offer a promising outlook for a field that is still relatively new. However, it is also imperative to remain vigilant regarding the prevention of data biases and under-representation in ML models to maximize the potential of these technologies.

\par Despite these promising developments, the field faces significant hurdles, notably the lack of readily available ``big data" in the medical domain. For instance, the routine digitization of histopathology slides remains a challenging goal in many healthcare facilities. Data sharing among medical institutions is fraught with challenges around appropriate procedures for the responsible sharing of patient data under institutional, national and international patient privacy regulations.

\par Advancing the field will likely entail overcoming these hurdles, ensuring more extensive sharing of de-identified data from research publications and greater participation in establishment of standardized public repositories for data. Dissemination of both code and pretrained model weights would also enable greater knowledge-sharing and repeatability. Models that incorporate uncertainty quantification, explainability, and strategies to account for missing data are particularly advantageous. For more guidance on building appropriate multimodal AI models in healthcare, one can refer to the World Health Organization's new ethics and governance guidelines for large multimodal models \cite{who2024ethics}.

\par In conclusion, the field of multimodal machine learning in biomedicine has experienced rapid growth in each of its challenge areas of representation, fusion, translation, alignment, and co-learning. Given the recent advancements in deep learning models, escalating interest in multimodality, and the necessity for multimodal applications in healthcare, it is likely that the field will continue to mature and broaden its clinical applications. In this ever-evolving intersection of AI and healthcare, the imperative for responsible innovation resonates strongly. The future of multimodal machine learning in the biomedical sphere presents immense potential but also mandates a dedication to ethical principles encompassing data privacy, accountability, and transparent collaboration between human professionals and AI systems. As we navigate this transformative journey, the collective effort, ethical stewardship, and adherence to best practices will ensure the realization of the benefits of AI and multimodal machine learning, making healthcare more efficient, accurate, and accessible, all while safeguarding the well-being of patients and upholding the procedural and ethical standards of clinical practice.


\begin{sidewaystable}[h]
\label{tab:table1}
\begin{center}
\begin{minipage}{\textwidth}
\caption{Literature relating to the five challenges of multimodal machine learning by the datatype analyzed.}\label{tab2}
\begin{tabular*}{\textwidth}{@{\extracolsep{\fill}}lcccc@{\extracolsep{\fill}}}
\toprule
& \multicolumn{4}{@{}c@{}}{Datatype}
\\\cmidrule{2-5}%
Challenge & MRI & CT & PET & EHR\\
\midrule
Representation    & \cite{Hamghalam2021,Zhang2022}   & \cite{Daza2020,Zhou2023}  &  & \cite{Daza2020,vanSonsbeek2020,Zhang2023,Wang2023b}\\

Fusion    & \cite{azcona2020interpretation, Neubauer2020, zhou2020m2net,Wang2021trans,Zhang2022,Rudie2022,Liu2023,Zhang2023,Zhou2023} & \cite{Daza2020,Neubauer2020,Yang2020,Bhalodia2021,LloretCarbonell2023}  & \cite{Neubauer2020} & \cite{Daza2020, Yang2020, vanSonsbeek2020,zhou2020m2net, Vivar2020, Li2021,Bhalodia2021,Cui2022,Khosravi2022}\\

Translation    & \cite{jiang2020unified, Hu2020,guo2020lesion, shin2020gandalf,Takagi2023}   & \cite{jiang2020unified,Zhu2020}  & \cite{Hu2020, shin2020gandalf} & \\

Alignment    & \cite{Wang2021trans,Zhang2022} & \cite{Yang2020,Leroy2023,Zhou2023,Li2023}  & & \cite{Yang2020,Li2023}\\

Co-learning    & \cite{varsavsky2020testtime, Yang2020b, Hu2020, Bui2020, Hu2021, Pei2023, Jafari2022, Xue2020, Dong2022}   & \cite{Xue2020,Hu2021, Pei2023, Jafari2022, Dong2022}  & & \cite{Xing2022} \\
\botrule
\end{tabular*}

\begin{tabular*}{\textwidth}{@{\extracolsep{\fill}}lccccc@{\extracolsep{\fill}}}
\toprule
& \multicolumn{5}{@{}c@{}}{Datatype}
\\\cmidrule{2-6}%
Challenge & Hist. & Ultrasound & Genomic & X-Ray & Fundus\\
\midrule
Representation & & & & \cite{vanSonsbeek2020}&\cite{Zhou2023}\\

Fusion  & \cite{Chen2020,Li2021,Cui2022}  & \cite{habib2020automatic} & \cite{Chen2020,Cui2022} & \cite{vanSonsbeek2020,habib2020automatic,Cui2022,LloretCarbonell2023,Khosravi2022}&\cite{Zhou2023}\\

Translation & & & & & \\

Alignment   & \cite{Leroy2023} & & & &\cite{Zhou2023}\\

Co-learning & & \cite{Xiong2020} & \cite{chauhan2020joint}& & \\
\botrule
\end{tabular*}

\end{minipage}
\end{center}
\end{sidewaystable}

\bibliography{references}

\backmatter
\bmhead{Author Contributions}
E.W. contributed the main writing of the paper. This paper concept was formulated by W.H., T.S.M., O.G., and J.L., A.R., W.H., T.S.M., C.E.K., and A.R. contributed ideas and direction for the writing and assisted in the proofreading and selection of the concepts and papers covered. E.W and A.R are greateful for support from NIH grant R37CA214955-01A1. All authors are grateful to support from the AMIA Biomedical Image Informatics Working group.

\bmhead{Data Availability}
No data outside of those referenced has been used in this survey. Key papers have been summarized in Table 1.

\bmhead{Competing Interests}
The Authors declare no competing financial interests but the following competing non-financial interests: A.R. serves as a member for Voxel Analytics, LLC. C.E.K.’s institution receives salary support for service as editor of Radiology: Artificial Intelligence.

\end{document}